\documentclass[fleqn,10pt]{wlscirep}
\usepackage[utf8]{inputenc}
\usepackage[T1]{fontenc}

\usepackage{mathtools}
\usepackage{algorithm}
\usepackage{algpseudocode}
\usepackage{ulem}

\usepackage{tikz}
\usetikzlibrary{arrows.meta, calc, positioning, decorations.pathreplacing}

\newcommand{\tk}{\textcolor{black}}
\newcommand{\tm}{\textcolor{black}}
\newcommand{\tx}{\textcolor{black}}
\newcommand{\ns}{\textcolor{black}}

\newcommand{\sn}{\textcolor{black}}

\title{\tk{SAVER: Stochastic Adaptive Variance-Driven Exploration {and Reconstruction }for Low-Dose Computed Tomography}}

\author[1]{Shunta Nonaga}
\author[2,3,*]{Koji Tabata}
\author[4,5]{Junya Honda}
\author[6,7]{Hiroyuki Kudo}
\author[7,8,9,10]{Wataru Yashiro}
\author[1,2,3,*]{Tamiki Komatsuzaki}
\affil[1]{Graduate School of Chemical Sciences and Engineering, Hokkaido University, Kita 13 Nishi 8, Kita-ku, Sapporo, Hokkaido 060-8628, Japan.}
\affil[2]{Research Center of Mathematics for Social Creativity, Research Institute for Electronic
Science, Hokkaido University, Kita 20 Nishi 10, Kita-ku, Sapporo, Hokkaido 001-0020, Japan}
\affil[3]{Institute for Chemical
Reaction Design and Discovery (WPI-ICReDD), Hokkaido University, Kita 21 Nishi 10, Kita-ku, Sapporo, Hokkaido
001-0021, Japan.}
\affil[4]{Graduate School of Informatics, Kyoto University, Yoshida-honmachi, Sakyo-ku, Kyoto 606-8501, Japan.}
\affil[5]{RIKEN Center for Advanced Intelligence Project (AIP), 1-4-1 Nihonbashi, Chuo-ku, Tokyo 103-0027, Japan.}
\affil[6]{Department of Computer Science, Institute of Systems and Information Engineering, University of Tsukuba, Tsukuba 305-8573, Japan.}
\affil[7]{International Center for Synchrotron Radiation Innovation Smart (SRIS), Tohoku University, Sendai, Miyagi 980-8577, Japan.}
\affil[8]{Institute of Multidisciplinary Research for Advanced Materials (IMRAM), Tohoku University, Sendai, Miyagi 980-8577, Japan.}
\affil[9]{Department of Finemechanics, Graduate School of Engineering, Tohoku University, Sendai, Miyagi 980-8579, Japan.}
\affil[10]{Graduate School of Dentistry, Tohoku University, Sendai, Miyagi 980-8575, Japan.}

\affil[*]{\tk{correspondence: tamiki@es.hokudai.ac.jp or ktabata@es.hokudai.ac.jp}}


\begin{abstract}
Computed Tomography (CT) is indispensable in clinical diagnostics, yet minimizing radiation dose \tk{without compromising image quality} remains a critical challenge. Conventional low-dose protocols often rely on fixed, uniform angular sampling, \tk{independent of the underlying structural complexity of organs} of individual patients. We propose ``Stochastic Adaptive Variance-Driven Exploration and Reconstruction'' (SAVER), an adaptive \ns{data} acquisition framework that selects projection angles in real-time based on the statistical variance of acquired data. Utilizing a Softmax-based stochastic scheduling \tk{scheme} with simulated annealing, SAVER prioritizes directions with high structural information while maintaining necessary exploration. Numerical experiments across 8 diverse phantoms demonstrate that SAVER achieves consistently higher reconstruction fidelity than conventional random sampling, particularly for objects with high structural anisotropy. Furthermore, the proposed method exhibits robust performance under significant measurement noise. By dynamically reallocating radiation dose to the most informative projections, SAVER provides a mathematically-grounded approach to maximize diagnostic quality per unit of radiation dose, marking a shift toward \ns{\tx{sample}-dependent}, data-driven CT acquisition.
\end{abstract}

\begin{document}

\flushbottom
\maketitle
%
%
\thispagestyle{empty}




\section*{Introduction}

Computed tomography (CT), first introduced by Hounsfield in the early 1970s, revolutionized diagnostic radiology by enabling non-invasive visualization of internal anatom\tk{y}~\cite{hounsfield1973computerized}. Over the past decades, continuous technological advances from early single-slice scanners to modern multi-slice and dual-energy CT systems have significantly improved spatial resolution, acquisition speed, and clinical applicability~\cite{brenner2007computed}. As a result, CT has become one of the most widely used imaging modalities for diagnosing a broad range of diseases. However, the growing use of CT has also raised concerns about the cumulative risk associated with ionizing radiation exposure. Consequently, reducing radiation dose while maintaining diagnostic image quality has become a central objective in CT research, following the principle of “as low as reasonably achievable” (ALARA)~\cite{valentin20082007}.

To address this issue, extensive research has focused on both hardware and computational approaches. Hardware-based strategies include beam filtration and detector improvements, while software-based approaches primarily rely on advanced image reconstruction techniques. In particular, iterative reconstruction (IR)~\cite{beister2012iterative,willemink2013iterative} and compressed sensing (CS)~\cite{caiafa2013multidimensional,kudo2013image} methods have demonstrated strong capabilities in reconstructing high-quality images from sparse-view or limited-angle measurements. These approaches have significantly improved the feasibility of low-dose CT~\cite{chen2017low,wolterink2017generative}, \tk{when the underlying assumption holds such as sparsity in the domain to be monitored for a given sample.} 
\tk{However, including IR and CS approaches, most existing CT acquisition strategies pre-design how to irradiate X ray to a sample in question irrespective of underlying structural complexity of the sample, and do not {explicitly} optimize the irradiation pattern, e.g., number of rays per angle, on the fly during the measurements with fewer a priori assumption on the sample.}

\begin{figure}[t]
\centering
\includegraphics[width=0.6\linewidth]{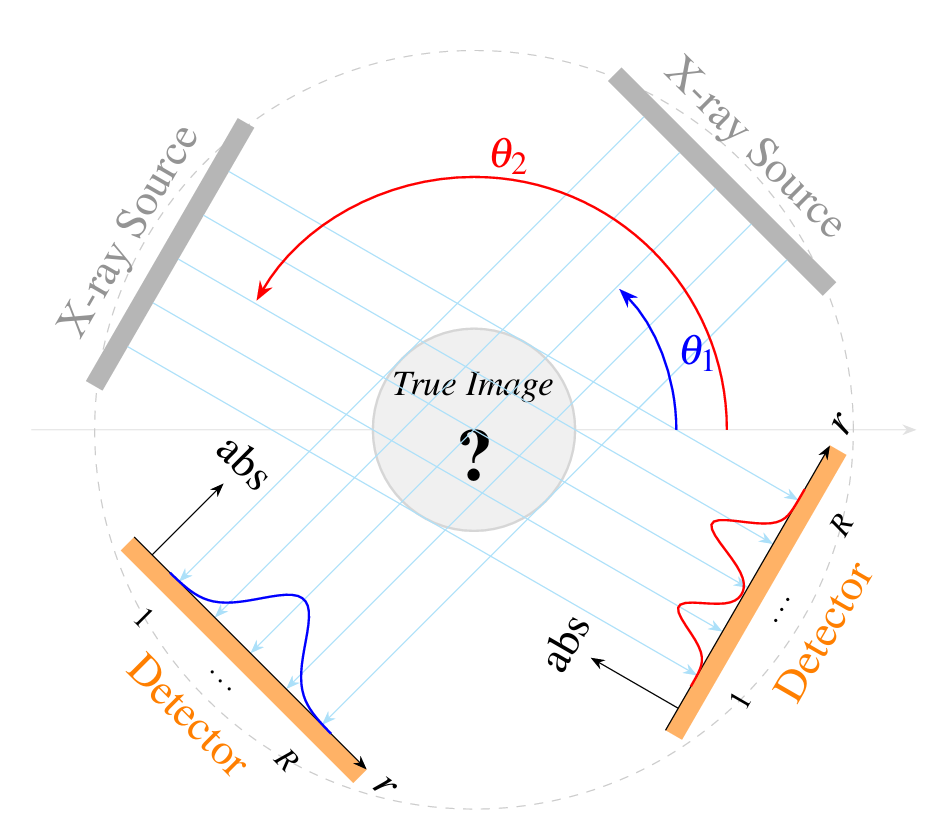}
    \caption{\textbf{Schematic illustration of the parallel-beam CT data acquisition process.} X-rays emitted from the source penetrate \tx{an} unknown central object (True Image) and are recorded by the detector as the absorption profile at each projection angle \tx{$\theta_{i}$}. The captured data at detector position $r$ reflect the structural complexity of the object; for instance, the profile at $\theta_1$ shows a simple unimodal distribution, whereas the profile at $\theta_2$ reveals more intricate internal features through multiple peaks in the absorption values.}
\label{fig:parallel}
\end{figure}

\tk{Uniform} angular sampling \tk{of X-ray irradiation means that the buried 2D (3D) structure of a sample is constructed from a set of X-ray absorption distributions resulting from uniform irradiation equally for different angles. This corresponds to implicitly assuming} that structural \tk{information (complexity)} is evenly distributed across projection angles \tk{(e.g., circle or sphere)}. In practice, however, anatomical structures often exhibit \tk{significant} anisotropy and spatial heterogeneity \ns{(See Fig.~\ref{fig:parallel})}. As a result, certain projection angles contain significantly more structural information than others, while many measurements contribute only marginally to reconstruction quality. From an information-efficiency perspective, this suggests that conventional fixed-geometry scanning may allocate radiation dose inefficiently by acquiring redundant projections that provide limited new information~\cite{dahmen2016feature,andersen2012statistical}.

Motivated by \tk{this}, we propose an adaptive CT acquisition framework that treats projection selection as a sequential decision problem, {where the system selects and acquires a single projection ray at each round}. The proposed method prioritizes projection \tk{angles} based on the \tk{sample} variance of the {observed \ns{projection data value}}, which serves as a proxy for structural complexity along each direction in Radon space. To balance global exploration of possible directions with focused sampling of informative ones, we introduce a stochastic scheduling strategy based on a Softmax policy with simulated annealing~\cite{kirkpatrick1983optimization,sutton1998reinforcement,jang2016categorical}. By dynamically allocating measurements toward projections with higher structural information, the proposed framework aims to improve reconstruction efficiency and maximize image quality per unit radiation dose.

\section*{Results and Discussion}
\subsection*{Problem Setup}

\begin{figure}[t]
\centering
\includegraphics[width=0.5\linewidth]{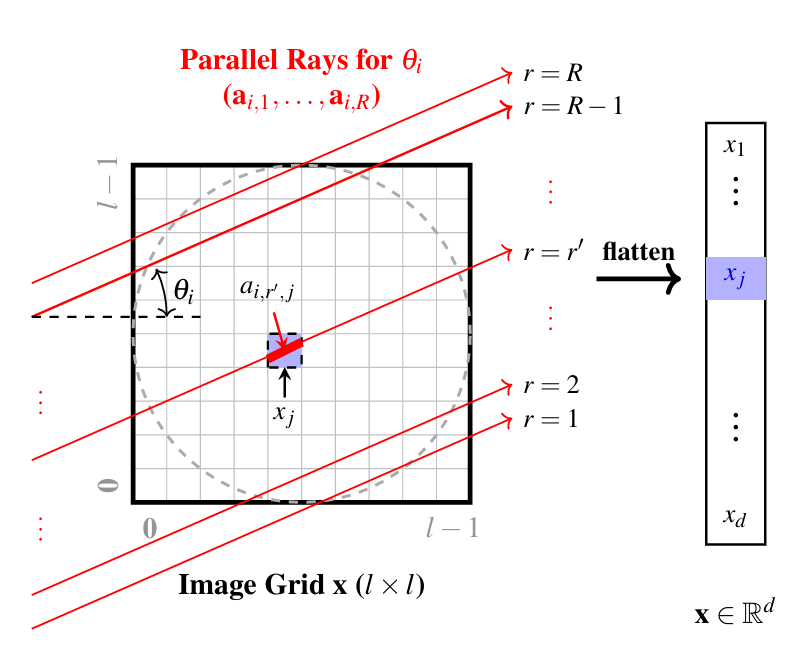}
    \caption{\textbf{Schematic \tx{illustration} of the CT forward model.} At each projection angle $\theta_i$, a set of parallel X-ray beams $r = 1, 2, \dots, R$ scans the $l \times l$ discretized image grid. The geometric interaction is represented by the intersection length $a_{i,r,j}$ between the $r$-th ray and the $j$-th pixel. For illustration, the segment of ray $r=r'$ passing through pixel $x_j$ is highlighted ($a_{i,r',j}$).}
    \label{fig:ct_model}
\end{figure}

In this study, the image reconstruction problem in CT is formulated as a linear inverse problem in which \tx{an} unknown image vector is estimated from sequential projection measurements. \ns{The discrete representation of the imaging geometry and the construction of the corresponding linear system are illustrated in Figure~\ref{fig:ct_model}.}

Let the unknown image be represented by a vector $\mathbf{x} \in \mathbb{R}^{d}$, where $d$ denotes the total number of pixels in the image. {Specifically, the continuous attenuation coefficient distribution is discretized into a finite grid of pixels, and each element of $\mathbf{x}$ represents the attenuation coefficient associated with a single pixel.} The set of available projection angles is defined as $\Theta=\{\theta_i=i\Delta\theta|i=0,1,\ldots,K-1\}$, where $\Delta\theta$ denotes the angular step size in degrees and $K = \left\lfloor \frac{180}{\Delta\theta} \right\rfloor$ is the total number of discrete angles in the range $[0,180)$. {\tk{Given} $[K] \coloneqq \{0, 1, \dots, K-1\}$, for each angle index $i \in [K]$, let $R$ denote the number of available irradiation positions. For the $r$-th irradiation position ($r \in \{1, 2, \dots, R\}$), we define a vector $\mathbf{a}_{i,r}$. The $j$-th element of $\mathbf{a}_{i,r}$ represents the intersection length between the $r$-th ray at angle $i$ and the $j$-th pixel ($j \in \{1, 2, \dots, d\}$).
}These vectors are computed using Joseph's \ns{reprojection method}~\cite{joseph2007improved}. 

In the proposed framework, {a projection measurement (referred to as a \ns{projection data value}) is acquired sequentially, which represents the line integral of the single projection ray attenuation coefficients along a specific ray path through the object in each acquisition round.}
\tk{(For proof-of-concept simulation experiment, we imposed a single projection ray rather than predetermined, uniformly distributed multiple rays irrespective of angles per round)}. At each round $t = 1, 2, 3, \ldots$, the algorithm selects an angle index $i_t$. Then, \tk{an} 
irradiation position \textcolor{black}{$r_t$ is randomly selected} \tk{from the unused position candidates up to round $t$}. The {observed \ns{projection data value}} $y_t$ is \tk{defined} as:
\begin{equation}\label{eq:projection}
y_t = \mathbf{a}_{i_t,r_t}^{\top} \mathbf{x} + \epsilon_t,
\end{equation}
where \textcolor{black}{for the purposes of our 
simulations, we assume} $\epsilon_t \sim \mathcal{N}(0,\sigma_B^2)$ to be independent Gaussian measurement noise with variance $\sigma_B^2$. After rounds of $t$, the observation vector becomes $\mathbf{y}_t=(y_1,y_2,\ldots,y_t)^\top\in\mathbb{R}^t$.
Let $A_t \coloneqq \left(\mathbf{a}_{i_1,r_1}, \mathbf{a}_{i_2,r_2},\ldots,\mathbf{a}_{i_t,r_t}\right)^\top\in\mathbb{R}^{t\times d}$ denote the corresponding measurement matrix. \tx{Here, $\mathbf{y}_t$ that would be acquired at the time when {\it all} irradiationable points are exhausted over all angles including less important angles corresponds to the set of absorption profiles at all projection angle in the parallel irradiation beam scheme shown in Figure~\ref{fig:parallel}. }
 The image estimate $\hat{\mathbf{x}}_t$ is obtained as the solution of the following minimization problem based on Tikhonov regularization~\cite{tikhonov1943stability}: $\hat{\mathbf{x}}_t \coloneqq \operatorname{argmin}_{\mathbf{x}\in\mathbb{R}^d}\ \left(\|A_t\mathbf{x} - \mathbf{y}_t\|_2^2 + \frac{\sigma_B^2}{\xi^2}\|\tk{\mathbf{x}}\|_2^2\right)$. The parameter $\xi^2$ controls the variance of the Gaussian prior distribution imposed on the unknown image ${\mathbf x}$ and serves as a hyperparameter. The stable estimate can be computed via \ns{solving} the normal equation as 
\begin{equation}\label{eq:normal eq}
    \hat{\mathbf{x}}_t=\left(A_t^\top A_t + \frac{\sigma_B^2}{\xi^2} I \right)^{-1}A_t^\top \mathbf{y}_t,
\end{equation}
where $I$ denotes the identity matrix. In this study, we aim to develop an adaptive control algorithm that computes an estimate $\hat{\mathbf{x}}$ close to the true image \tk{$\mathbf{x}$} while using as fewer X-ray projections as possible.

\subsection*{Proposed Framework: SAVER}

To address 
sampling inefficiency, we introduce Stochastic Adaptive Variance-Driven Exploration and Reconstruction (SAVER). The core idea is to treat the CT acquisition as a sequential decision process where the system dynamically reallocates radiation dose based on 
structural complexity of the \tk{sample} object. \tk{The working hypothesis is that the higher the variance in \ns{projection data values} (associated with the corresponding X-ray absorption distribution) along a specific projection angle, the more the structural information of the sample object is concentrated in reconstructing the image $\mathbf{x}$.}
{SAVER prioritizes increasing the number of rays for such informative directions to maximize reconstruction fidelity per dose, ensuring that each specific projection ray is sampled without repetition.}

\tk{
Let $Y_i(t)$ be the set of \ns{projection data values} obtained from angle index $i$ up to round $t$, with the total number of the observations $n_i(t) = |Y_i(t)|$.
The sample variance for index $i$ is computed as 
\begin{equation}\label{eq:samplevariance}
\hat{\sigma}_i^2(t)
\coloneqq
\frac{1}{n_i(t)} \sum_{y \in Y_i(t)} y^2
-\left(
\frac{1}{n_i(t)} \sum_{y \in Y_i(t)} y
\right)^2.
\end{equation}
The sample variance $\hat{\sigma}_i^2(t)$ quantifies structural variability observed from index $i$ and is updated under the process of measurements. Especially when the values of $n_i(t)$ are small, $\hat{\sigma}_i^2(t)$ can vary abruptly across angles, and thus we employ robust scaling based on the median and the interquartile range (IQR), insensitive to outliers, defined by $\tilde{\sigma}_i(t)
=\frac{
\hat{\sigma}_i^2(t) - \mathrm{median}(\mathbf{v}(t))}{
\mathrm{IQR}(\mathbf{v}(t))
}$, where $\mathbf{v}(t)$ is a set of $\hat{\sigma}_i^2(t)$ whose total number of points that have been irradiated up to round $t$ is smaller than the maximum and $\mathrm{IQR}(\mathbf{v}(t))$ denotes the interquartile range of $\mathbf{v}(t)$. The normalized value $\tilde{\sigma}_i(t)$ is then used as the score for the Softmax-based angle selection policy. The next projection angle is selected probabilistically via a ``temperature''-controlled Softmax function: the selection probability of projection angle $i$ at round $t$ over an angle set (denoted by $F(t)$) whose total number of points that have been irradiated up to round $t$ is smaller than the allowance is defined by
\begin{equation}
P_i(t)
\coloneqq
\frac{
\exp\!\left(\frac{\tilde{\sigma}_i(t)}{T_t}
\right)
}{
\sum_{j\in F(t)}
\exp\!\left(
\frac{\tilde{\sigma}_j(t)}{T_t}
\right)
}.
\label{eq:softmax}
\end{equation}
$T_t (> 0)$ denotes the ``temperature'' parameter to control sampling over different angles, i.e., when $T_t$ is larger (smaller), the probability distribution becomes more (less) uniform, promoting exploratory sampling across projection angles (exploitation of informative orientations). In this paper,
to gradually shift the sampling behavior from exploration to exploitation, $T_t$ is updated by $T_t=\max
\left(
T_{\min},
\frac{T_{\text{initial}}}{1 + \eta (t-2K)}
\right)$, where $T_{\text{initial}}$ ($T_{\min}$) denotes the initial temperature (the minimum temperature), and $\eta$ represents the annealing rate. This annealing strategy ensures that the algorithm initially explores a wide range of projection angles and progressively concentrates measurements on orientations that exhibit higher structural variability.
}


Unlike conventional fixed geometries, SAVER utilizes a probabilistic scheduling policy that balances the exploration of unknown projection angles with the exploitation of highly informative directions identified through real-time variance analysis. 
\tx{The overall operational workflow of the proposed method composed of the two-stage process (initialization and adaptive scheduling) is illustrated in Figure~\ref{fig:proposed} and the role of each hyperparameter is summarized in Table~\ref{tab:parameters} in the Methods section. It should be noted that SAVER does not necessarily require the inverse matrix calculation $B_t^{-1}$ and restore the updated image $\hat{\mathbf{x}}_t$ at every round for adaptive decision-making. The image reconstruction process can be performed only when necessary, thereby significantly reducing the total computational overhead during the acquisition phase (In this paper, however,  we reconstructed and updated the underlying image $\hat{\mathbf{x}}_t$ using Woodbury Matrix Identity presented in the Methods section, for the purpose of elucidation of the reconstruction performance of SAVER and its variants, especially to compare with the conventional scheme (Random) without referring any variance information).}

\subsection*{Experimental Setup}
\begin{figure}[t]
\centering
\includegraphics[width=\linewidth]{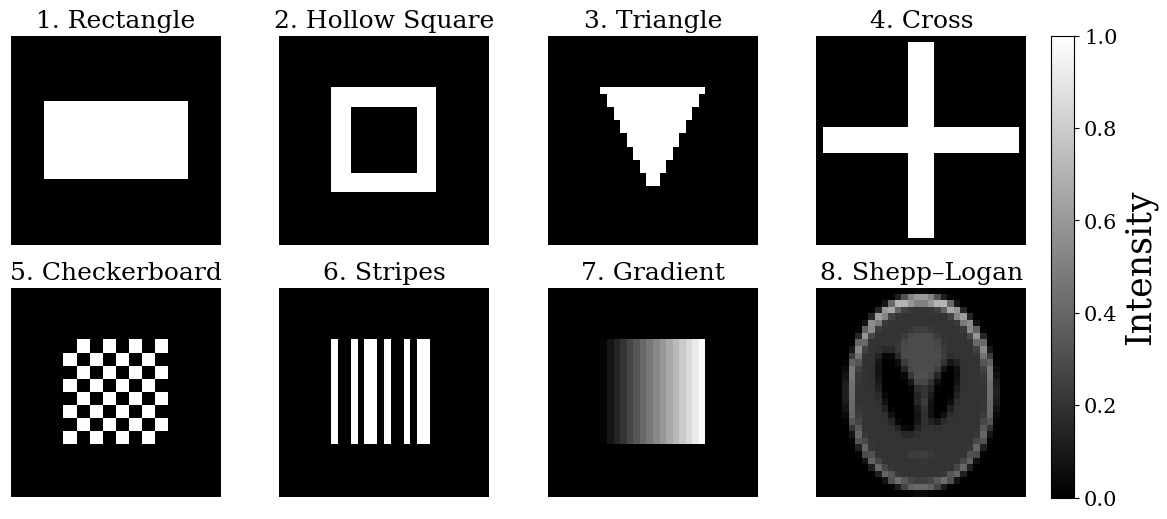}
\caption{{\textbf{8 synthetic phantoms used in the experiments}: (1) Rectangle, (2) Hollow Square, (3) Triangle, (4) Cross, (5)
Checkerboard, (6) Stripes, (7) Gradient, (8) Shepp-Logan~\cite{shepp1974fourier}. All images have a resolution of $32\times32$ pixels.}}
\label{fig:phantom}
\end{figure}

We conduct numerical simulations under a simplified CT imaging model assuming a monochromatic X-ray beam and a linear attenuation process. Under this model, each projection measurement corresponds to the line integral of the object’s attenuation coefficient, while physical effects such as beam hardening, scattering, and refraction are neglected.
To evaluate the proposed method, we perform experiments on 8 computational phantoms ($32\times32$ pixels; Figure~\ref{fig:phantom}) representing diverse structural characteristics. \sn{For all projection angles, the number of available irradiation positions was set to $R=32$, matching the image size.} {The performance of the proposed framework was evaluated against several adaptive and non-adaptive baselines. In addition to the original SAVER algorithm, we also consider three practical methods and four oracle-guided reference methods.
\begin{itemize}
    \item \textbf{Random:} Selects a projection angle $i_t$ and an irradiation position $r_t$ uniformly at random from the set of all available irradiation positions without replacement \tk{(=no irradiation is applied for the same position)}.
    \item \textbf{AIRS (Axis-Initialized Random Sampling):} 
    Before random sampling begins, AIRS first \tx{utilizes (exhausts) all} irradiation positions at $0^\circ$ and $90^\circ$, which correspond to the principal horizontal and vertical directions of the \tk{space-fixed} image grid 
    \tk{(how an object to be measured is lying in the grid is, of course, assumed to be unknown)}. After this deterministic initialization, the remaining irradiation positions are selected uniformly at random without replacement. 
    \item \textbf{SAVER-A:} 
    SAVER-A first exhausts the irradiation positions at $0^\circ$ and $90^\circ$. After this axis-prior initialization, the algorithm follows SAVER.
    \item \textbf{MAX-V (Oracle):} \tk{Selects a projection angle $i_t$ from the angle whose {\it true} variance is the largest, sequentially in descent order up to the smallest {\it true} variance angle, each of which all irradiation positions are sampled for the chosen angle via Round-Robin (Remind that unless irradiating the object in full from all angles, one cannot acquire {\it true} variances of the \ns{projection data values} for all angles).} 
    \item \textbf{MIN-V (Oracle):}  \tk{Selects a projection angle $i_t$ from the angle whose {\it true} variance is the smallest, sequentially  in ascent order up to the largest {\it true} variance angle, each of which all irradiation positions are sampled for the chosen angle via Round-Robin.}
    \item \textbf{SAVER-O (Oracle):}  \tk{Selects a projection angle $i_t$ according to Softmax probabilities using {\it true} variances of all angles. Projection angles whose irradiation positions have been fully consumed are excluded from the probability calculation. No explicit initialization phase is required.} 
    \item \textbf{SAVER-AO (Axis-Oracle):}  \tk{First exhausts the irradiation positions at $0^\circ$ and $90^\circ$ in the same manner as SAVER-A, and subsequently follows the same oracle-guided Softmax scheduling as SAVER-O.} 
\end{itemize}
}
The performance gap between MAX-V and MIN-V serves as a direct measure of the utility of the variance metric in prioritizing structural information. 

In our simulations, the total number of acquisition rounds \tk{$N$}
was set to $N=500$. \tk{Each acquisition round here corresponds to an X-ray irradiation of a certain position for a chosen angle. Note that without noise, $N\simeq 800$ is a sufficient number of irradiation points to reconstruct the underlying image \ns{accurately} 
in the present setting.} {
Each measurement is given by Eq.~\ref{eq:projection}. Stacking observations up to round $N$, we obtain a linear system $\mathbf{y}_N=A_N\mathbf{x}$, where $A_N\in\mathbb{R}^{N\times d}$ consists of row vectors. In the noiseless case, accurate reconstruction requires that this system is (approximately) \ns{inverted}, i.e., the number of linearly independent measurements is comparable to the number of effective unknowns. For an image of size $l\times l$, the total number of pixels is $d=l^2$. However, since reconstruction is restricted to the inscribed circular region, the effective number of unknowns is approximately $d\times\frac{\pi}{4}=l^2\times\frac{\pi}{4}$. For $l=32$, this yields $32^2\times\frac{\pi}{4}=804.2477\ldots$. Therefore, $N\simeq 800$ measurements are sufficient to reconstruct any object in the noiseless setting.} 

To investigate the robustness of the proposed framework against stochastic perturbations, two distinct measurement noise regimes were considered: a low-noise environment ($\sigma_B = 10^{-6}$), approximating a near-ideal scenario, and a high-noise environment ($\sigma_B = 10^{-2}$), {which may reflect low S/N cases for actual CT observation}. 
The configuration for the SAVER algorithm was standardized with an angular resolution of $\Delta\theta = 3$ \tk{degree} and a prior regularization parameter of $\xi = 0.1$. {To balance exploration and exploitation, we used a Softmax-based scheduling policy with temperature annealing (see Methods). The temperature parameter $\tk{T}_t$ controls the sharpness of the selection distribution and is gradually decreased to shift the policy from exploration to exploitation. In our experiments, $\tk{T}_t$ was initialized at $\tk{T}_{\text{initial}} = 1$ and decayed to $\tk{T}_{\min} = 0.1$ with annealing rates $\eta \in \{1, 0.1, 0.01\}$.}
%
For each experimental configuration, the reconstruction process was repeated ten times using independent random seeds to ensure statistical significance. {To ensure that the performance gains are invariant to the object's orientation relative to the Cartesian image grid and to avoid potential grid bias, the phantom image in each of the ten independent trials was rotated by a random integer angle $\alpha\in[0,45]^\circ$ prior to the acquisition process. This randomization ensures that the adaptive selection policy is evaluated on its ability to identify structural complexity regardless of how the object is aligned with the pixel grid.} The reconstruction fidelity at each acquisition round $t$ was quantified using the mean Structural Similarity Index Measure (SSIM) and its associated standard deviation, computed relative to the ground truth image within the inscribed circular field of view as defined in Eq.~\ref{eq:ssim}. {In the following sections, we primarily discuss the results obtained with the annealing rate $\eta=0.01$ under a low-noise environment ($\sigma_B = 10^{-6}$). The results and discussion for other annealing rates ($\eta=\tx{0.1,1,10}$) under both noise environments are provided in the Supplementary Information for comparative analysis.}

\section*{Discussion}

\tk{
To begin, we show a sinogram representation of the 8 synthetic phantoms in Figure~\ref{fig:sinogram}. They are graphical encodings of X-ray projections collected at multiple angles around objects. Each of them captures how X-ray intensity varies as the beam passes through the object from different directions. In this representation, the horizontal axis corresponds to detector position, while the vertical axis represents the projection angle. Sinogram contains not only the information needed to infer internal structures for each angle, and also importantly how inhomogeneous the information each angle carries is. One can see immediately that some phantoms such as {\it 1. Rectangle}, {exhibit significantly different intensities (integrated over positions) depending on the angle, whereas other phantoms} such as {\it 8. Shepp-Logan}, carry {relatively uniform intensities across all angles, as expected from the phantom images.}}

\tk{Then, {\it How much does the working hypothesis actually work in reconstructing the image $\mathbf{x}$ based on the variances in \ns{projection data values} (associated with the corresponding X-ray absorption distribution) along projection angles?}  Figure \ref{fig:results_low} 
presents the evolution of the SSIM as a function of the number of acquisition rounds under low-noise ($\sigma_B = 10^{-6}$) for the 8 phantoms.
\begin{figure}[t]
\centering
\includegraphics[width=\linewidth]{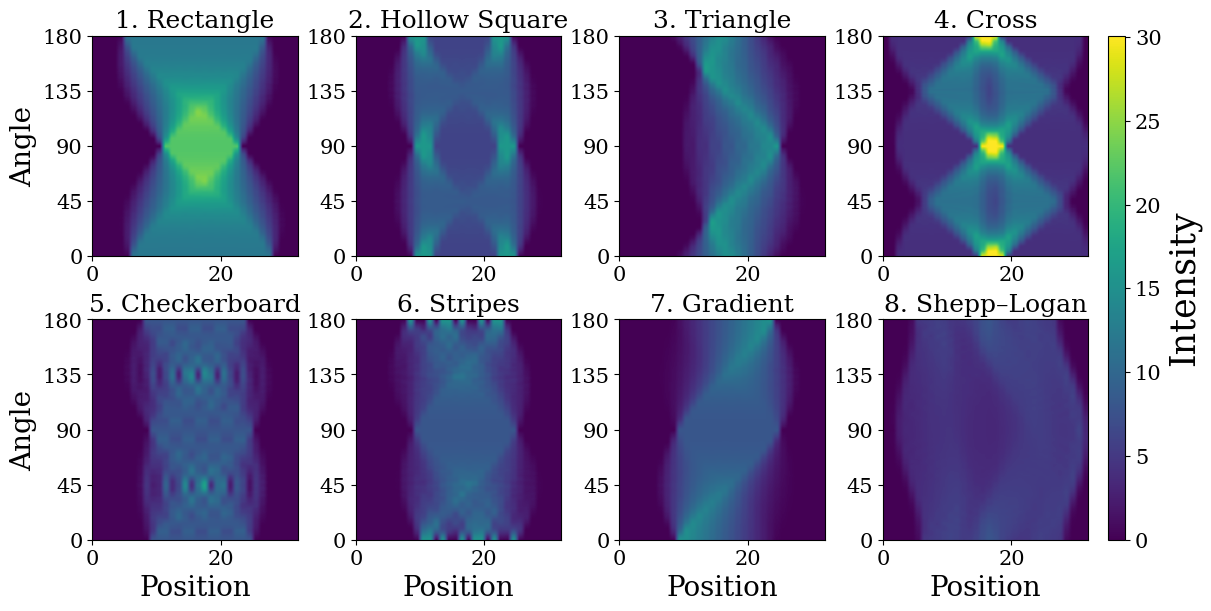}
\caption{\textbf{Sinogram representations of the 8 synthetic phantoms {in their original orientations (i.e., without the random rotation $\alpha$ used in the experiments)}.} The horizontal axis represents the \tx{radial} position (detector index), and the vertical axis denotes the projection angle in degrees, ranging from $0^\circ$ to $180^\circ$. These sinograms illustrate the distinct structural signatures in the Radon space that drive the variance-based adaptive selection process.}
\label{fig:sinogram}
\end{figure}
\begin{figure}[t]
\centering
\includegraphics[width=\linewidth]{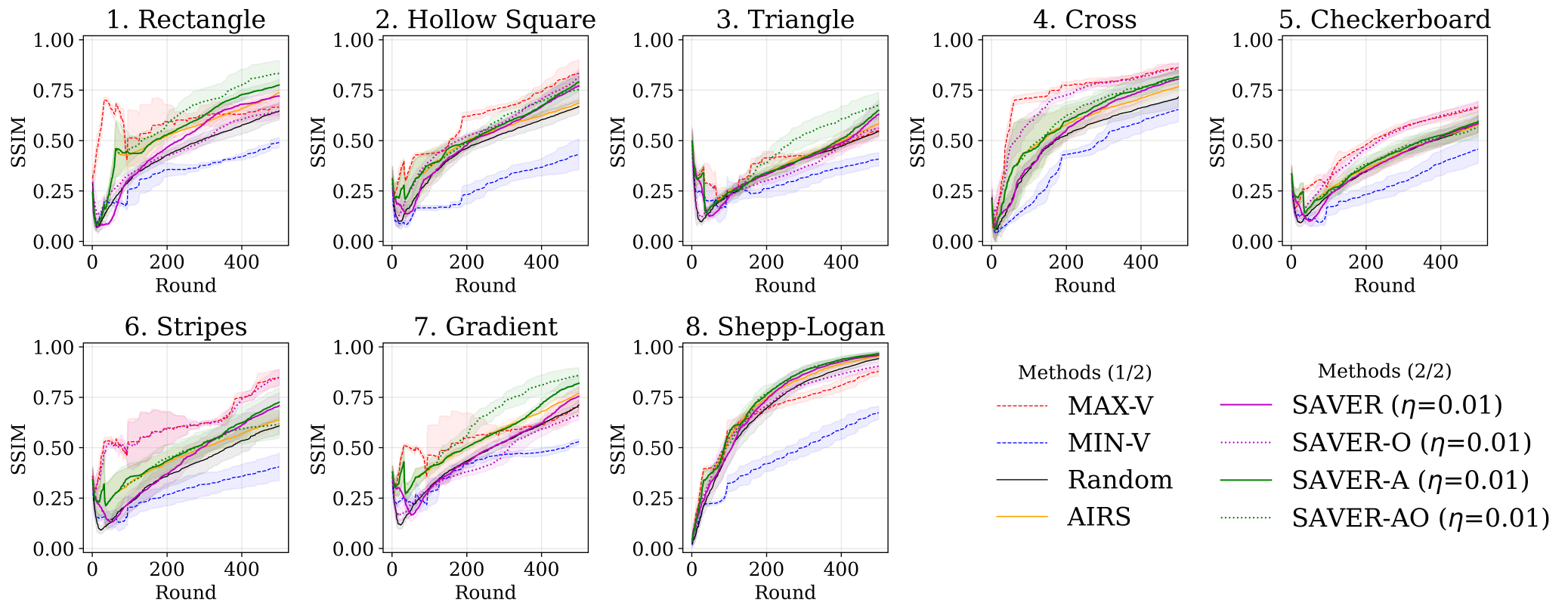}
\caption{\textbf{Reconstruction performance as a function of acquisition rounds in the low-noise environment} ($\sigma_B = 10^{-6}$) \textbf{for the 8 phantoms}. {The horizontal axis (Round) denotes the cumulative number of X-ray measurements, and the vertical axis represents the SSIM relative to the ground truth. For all SAVER variants, the results are shown for the annealing rate $\eta = 0.01$. Solid lines indicate the practical methods (SAVER, SAVER-A) and non-adaptive baselines (Random, AIRS), while dashed and dotted lines represent the oracle-guided reference methods (MAX-V, MIN-V, SAVER-O, SAVER-AO). Shaded regions denote the standard deviation over ten independent trials.}}
\label{fig:results_low}
\end{figure}
\begin{figure}[t]
\centering
\includegraphics[width=\linewidth]{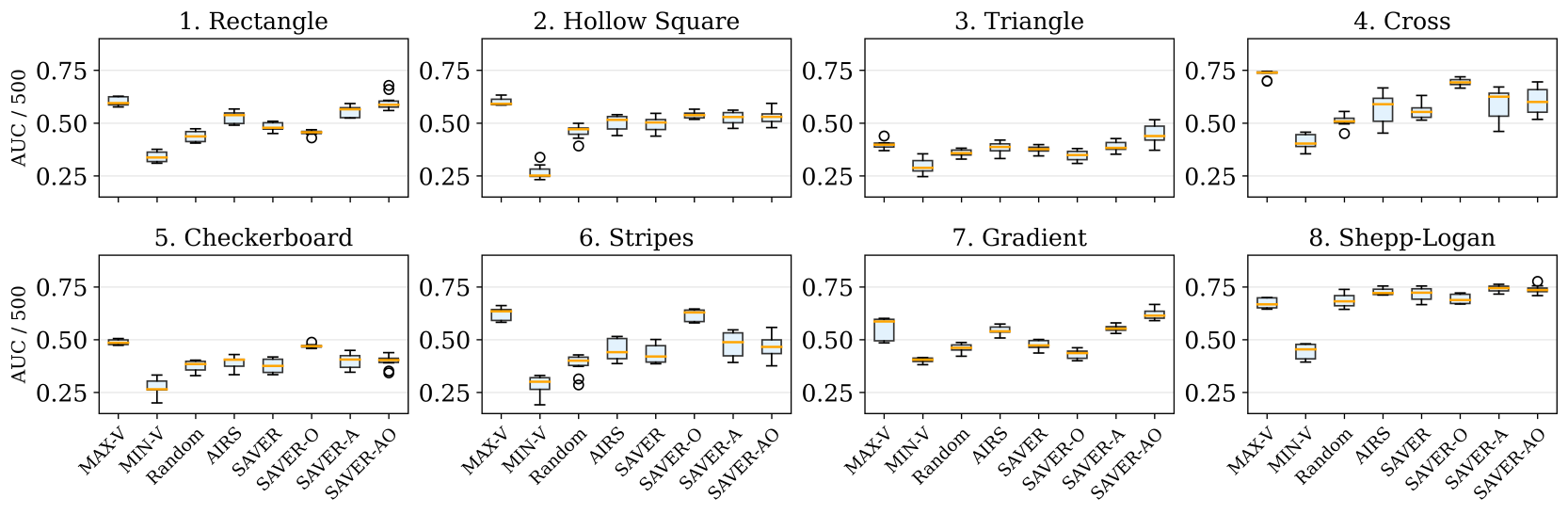}
\caption{{\textbf{Comparison of normalized Area Under the Curve (AUC/500) for SSIM across 8 phantoms in the low-noise environment} ($\sigma_B=10^{-6}$). Each box plot represents the distribution of values over 10 independent trials. The AUC is calculated by integrating the SSIM curve and dividing by the total number of rounds ($N=500$). Higher values indicate superior cumulative reconstruction fidelity throughout the acquisition process.}}
\label{fig:auc_low}
\end{figure}
\begin{figure}[t]
\centering
\includegraphics[width=\linewidth]{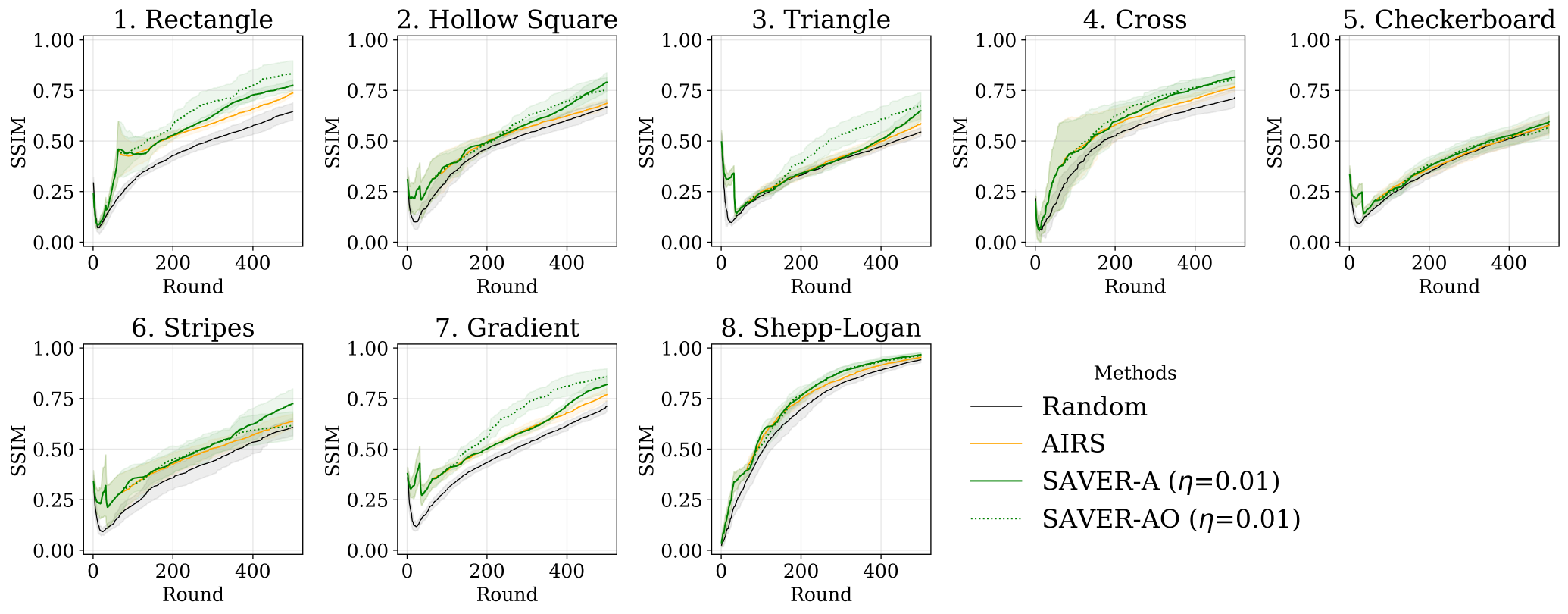}
\caption{{\textbf{Comparative analysis of four selected methods to evaluate the effectiveness of axis-prior initialization.} This figure highlights a subset of strategies from Figure~\ref{fig:results_low} to specifically compare non-adaptive and adaptive axis-initialized methods. The evolution of SSIM for Random, AIRS, SAVER-A, and SAVER-AO is shown for all 8 phantoms. By isolating these four methods, it becomes evident that real-time variance estimation (SAVER-A) effectively utilizes structural information that simple deterministic axis-sampling (AIRS) fails to capture.}}
\label{fig:axis}
\end{figure}
Let us first focus on the behaviors of the oracle methods, MAX-V (red dashed line) and MIN-V (blue dashed line) with the conventional scheme, Random (black solid line) without referring any variance information of the \ns{projection data values} of angles. MIN-V {generally exhibits the lowest SSIM performance across most rounds}
for all phantoms. 
This partially indicates that the lesser the variance of \ns{projection data values} for angle, the lesser the importance of angles in capturing the object image. 
In turn, MAX-V tends to yield a high performance, especially for phantoms in which the projection angles have inhomogeneous importance in sinogram  (Fig. \ref{fig:sinogram}) such as \tm{{\it 1. Rectangle}, 
{\it 4. Cross}, and {\it  6. Stripes}} resulting in a substantial gap between MAX-V and MIN-V, and the baseline algorithm Random is located somewhere between the two. MAX-V performs better than Random notably during early rounds for those phantoms.
\tm{MAX-V also exhibits a higher performance than Random in the reconstruction for phantoms {\it 2. Hollow Square} and {\it 5. Checkerboard} whose images (Fig. \ref{fig:phantom}) are somehow symmetric \tm{as well as {\it 7. Gradient.}}.}
 In turn, for the phantoms whose variance information along projection angles are more homogeneous, such as \tm{{\it 3. Triangle} and {\it 8. Shepp-Logan}}, 
MAX-V exhibits only comparative performance against Random.
However, it may be noted that during early rounds (say, less than 50 rounds) MAX-V shows slightly higher performance than Random for {these \tm{two} phantoms.}
All these observation suggests that, especially  for ``structured'' object where each projection angle carries uneven information for the underlying object, at least during early rounds,  nonrandom choice of projection angles are effectively in capturing the underlying objects.}

\tk{However, no one knows the underlying true variance of \ns{projection data values} along angles a priori.} {This fundamental limitation necessitates an acquisition framework capable of estimating structural complexity in real-time, purely from the data acquired during the measurement process. The SAVER algorithm addresses this challenge through its two-stage operational workflow (as illustrated in Fig.~\ref{fig:proposed}): a deterministic initialization phase to establish baseline variance estimates, followed by an adaptive scheduling phase that probabilistically steers irradiation toward highly informative directions. By bridging the gap between unknown structural distributions and data-driven decision-making, SAVER transforms CT from a static process into a dynamic, sample-dependent exploration. The following subsections analyze how this mechanism effectively manages information scarcity and adapts to various sample geometries.}

\subsection*{Information Scarcity and Algorithmic Efficiency}
{The efficacy of the SAVER framework is fundamentally rooted in its ability to exploit the structural anisotropy of the sample. As visually evident in the sinograms (Fig.~\ref{fig:sinogram}), structural information is concentrated in narrow angular bands for objects with high geometric non-uniformity, such as \textit{1. Rectangle}, \tm{{\it 4. Cross}}, and \textit{6. Stripes}. \tk{
In Figure.~\ref{fig:results_low}, for \textit{1. Rectangle}, \tm{{\it 4. Cross}}, and \textit{6. Stripes} phantoms, SAVER-A (green solid line) reaches the target fidelity significantly faster than conventional uniform sampling (Random) (black solid line), which does not account for structural heterogeneity. \tm{Note that SAVER-A also reach the target fidelity (slightly) faster than Random in {\it 2. Hollow Square} and {\it 7. Gradient} ({\it 8. Shepp-Logan}).} While the SSIM convergence curves as a function of acquisition rounds (Fig.~\ref{fig:results_low}) provide an overall comparison of algorithm performance across phantoms, their intrinsic complexity—arising from overlapping standard deviation bands and fluctuating recovery behavior over 8 distinct phantoms—limits detailed quantitative interpretation.}}
\tk{To overcome these visual limitations, we introduce the normalized Area Under the Curve (AUC/500) for SSIM as a summary metric. {Note that this AUC/500 is the integral of the SSIM curve and is distinct from the AUC of a Receiver Operating Characteristic (ROC) curve.} By integrating the fidelity curve over the entire span of 500 acquisition rounds, the AUC provides a 
brief
summary of the reconstruction performance, that is, not only for the final image quality but also for how close to a hypothetical case in which only the first one round can reconstruct the underlying object (SSIM $\simeq 1$) and consecutive rounds up to 500 maintain that quality.
Figure~\ref{fig:auc_low} presents the box-whisker plots of AUC/500 for each algorithm with different phantoms. As seen in Figure~\ref{fig:auc_low}, MIN-V has the lowest AUC/500 values for most phantoms, and MAX-V {tends to yield higher} in AUC/500 {values} than the others \tm{for the phantoms with heterogeneous importance in projection angles such as \textit{1. Rectangle}, \textit{4. Cross}, and \textit{6. Stripes}, including \textit{2. Hollow Square} and \textit{5. Checkerboard}.} 
Note that AIRS improves the performance of Random especially for phantoms having heterogeneous preference in projection angles.  In comparison between SAVER variants (SAVER and SAVER-A), SAVER-A exhibits better or at least equal performance to SAVER.
}

\subsection*{Analysis of Axis-Prior Initialization and Structural Heterogeneity}

{
The core logic behind starting the acquisition with orthogonal projections ($0^\circ$ and $90^\circ$) \tk{employed in AIRS, SAVER-A, and SAVER-A} is rooted in early basis construction for the image estimate; a detailed mathematical interpretation regarding how this stabilizes the posterior covariance matrix and improves the conditioning of the sequential inverse problem is provided in \tm{the Supporting Information, Section 1.}}
\tk{Figure~\ref{fig:axis} presents SSIM as a function of acquisition rounds only for the three Axis-Prior Initialization-based algorithms in addition to Random corresponding to the conventional algorithm in Figure~\ref{fig:results_low}.}
{The performance gains observed in Figure~\ref{fig:axis} are directly dictated by the structural characteristics of the phantoms in Radon space. For phantoms exhibiting high structural anisotropy, such as {\it 1. Rectangle}, \tm{{\it 4. Cross}}, and {\it 6. stripes}, the sinograms in Figure~\ref{fig:sinogram} reveal that information is concentrated within very narrow angular bands. In these cases, Figure~\ref{fig:axis} shows statistically significant 
performance gaps between SAVER-A (green solid line) and AIRS (orange solid line) \tm{in the exploitation stage where the annealing has been performed}. While AIRS benefits from the initial stabilization of the axis-prior, its subsequent uniform random sampling fails to exploit the highly localized structural information. In contrast, SAVER-A utilizes real-time variance estimation to identify and concentrate radiation dose on the ``informative'' peaks seen in the sinogram, leading to a much steeper rise in SSIM during the early-to-mid acquisition rounds.}

{Conversely, for phantoms with more homogeneous or isotropic information distributions, such as {\it 8. Shepp-Logan}, the sinogram in Figure~\ref{fig:sinogram} shows relatively consistent intensity across the $0^\circ-180^\circ$ range. Consequently, as seen in Figure~\ref{fig:axis}, the performance gap between the four selected methods is narrower. In such cases, because most projection angles carry comparable amounts of information, the advantage of adaptive selection is reduced. However, it is noteworthy that SAVER-A still performs at or above the level of the AIRS and Random baselines, demonstrating that the stochastic scheduling policy does not ``over-invest'' in specific angles when the sample does not warrant it.}
\tm{In turn, for {\it 5. Checkerboard} having almost circular symmetric nature in shape more than {\it 8. Shepp-Logan}, most of all algorithms including SEVER-AO using the underlying true variance exhibit similar performance as non-adaptive baseline Random.}
{The comparison in Figure~\ref{fig:axis} also highlights the behavior of \tm{phantoms such as {\it 3. Triangle} and {\it 7. Gradient}, indicating that, if the estimation of the underlying true variance would be tighter (more accurately estimated) (see SAVER-AO), SAVER-A should outperform the baseline AIRS as those exhibiting high structural anisotropy.} }

Overall, 
we observe that AIRS plateaus shortly after the initial axis-exhaustion phase. SAVER-A, however, successfully navigates the exploration-exploitation trade-off to find these critical off-axis directions. This confirms that while the axis-prior provides a necessary ``warm-start'' (as discussed in the Supplementary Information), it is the subsequent adaptive variance-driven scheduling that allows the system to transcend fixed geometric assumptions and adapt to the specific orientation and complexity of the individual sample.

\subsection*{Stability via Stochastic Selection Policy}
{The stability of the SAVER framework stems from its \textit{Softmax-based stochastic policy}. Unlike deterministic greedy strategies that may over-invest in narrow angular ranges, the simulated annealing approach maintains a necessary degree of exploration. This results in the smooth and monotonic improvement of the SSIM observed across all phantoms in Figure~\ref{fig:results_low}. While the primary analysis focuses on this low-noise environment to elucidate the core principles of SAVER, the framework's robustness against significant measurement noise ($\sigma_B = 10^{-2}$) is also confirmed. Under such high-noise regimes, the stochastic policy acts as a vital regularizer, preventing the system from becoming ill-conditioned. The detailed comparative analysis with oracle-guided greedy methods under noise is provided in the Supplementary Information (Figs.~S1--S3).}


\section*{Outlook}

In summary, the efficiency of the SAVER framework is driven by its ability to adapt to the \textit{nonuniformity} of information in the sinogram (See Fig.~\ref{fig:sinogram}). While random sampling implicitly treats all projection angles as equally informative, SAVER dynamically reallocates radiation dose to the most critical ones. \tk{Overall,} {this study not only demonstrates the efficiency of adaptive sampling but also reveals that a stochastic policy is essential for maintaining stability under measurement noise, marking a significant step toward personalized, data-driven CT acquisition.}
\tk{The following are some of the problems to be uncovered in the forthcoming subjects in autonomous, data-driven design of irradiation profile in CT reconstruction.}

First, while the proposed SAVER framework adaptively selects projection angles, the irradiation position within each selected angle is currently sampled in a simple sequential manner. Investigating more principled irradiation position selection strategies is an important next step. In particular, quasi–Monte Carlo (QMC) sampling, such as Sobol or Halton sequences~\cite{joe2008constructing}, may provide a more uniform coverage of the detector coordinate and potentially improve reconstruction stability. Other space-filling designs, including Latin hypercube sampling~\cite{loh1996latin} and blue-noise sampling~\cite{parada2019blue}, \tk{may} also be explored to ensure more uniform and less correlated ray distributions. A systematic comparison of these sampling strategies \tk{should} clarify their impact on reconstruction quality and sampling efficiency.

Second, the proposed approach currently focuses on selecting projection angles based on variance-driven exploration, without explicitly optimizing the posterior covariance of the reconstruction. Future work \tk{may} incorporate principles from optimal experimental design to guide the selection of irradiation positions. In particular, D-optimality, which minimizes the determinant of the posterior covariance matrix, and A-optimality, which minimizes its trace, may provide principled criteria for sequentially selecting irradiation positions that maximally reduce reconstruction uncertainty~\cite{pukelsheim2006optimal}. Integrating such criteria with the SAVER framework \tk{is expected to} lead to more efficient information acquisition and improved reconstruction accuracy.

Third, the current Softmax-based policy provides a simple mechanism for balancing exploration and exploitation. However, more principled bandit-based policies \sn{(as frameworks for sequential decision-making under uncertainty)}~\cite{lattimore2020bandit} may yield stronger theoretical guarantees. In particular, upper confidence bound (UCB)~\cite{auer2002finite} strategies \tk{are among the subjects to be addressed}
for projection-angle selection. By deriving tighter confidence intervals for the variance using concentration inequalities such as the McDiarmid inequality~\cite{hou2022almost} or related bounds, it may be possible to design a theoretically grounded UCB policy tailored to the CT acquisition process.

\tk{Fourth,} 
irradiation position selection can be guided not only by reconstruction uncertainty \tk{(in the present study)} but also by downstream objectives such as lesion detectability or segmentation performance\tk{, which integrate task-driven acquisition strategies on multiobjective fashion. }
Combining the proposed adaptive acquisition strategy with modern learned reconstruction methods, including plug-and-play priors or diffusion-based reconstruction models, may \tk{also} further enhance reconstruction quality under extremely low-dose conditions.
\tk{Finally, the realization of our proposed framework to three-dimensional cone-beam CT~\cite{patel2015cone}, where the acquisition geometry introduces additional degrees of freedom and challenges, is one of the possible steps toward realization into real applications. }

\tm{Fifth, as seen in Fig.~\ref{fig:sinogram}, structural anisotropy exists not only angle axis but also position axis. In this paper, we focused only on the preferential selection of angles in reconstructing the underlying image but there exists room to optimize and design which positions are selected during the CT reconstruction. This is one of the on-going projects to be addressed in a separate paper.} 

\section*{Methods}

\subsection*{Summary of Designed Parameters}
\tx{The} key user-defined parameters and their roles are summarized in Table~\ref{tab:parameters}. These parameters \tx{control the balance between exploration of unknown projection angles and exploitation of information acquired up to each round.} 

 \begin{table}[ht]
 \centering
 \caption{Summary of key parameters in the SAVER framework.}
 \label{tab:parameters}
 \begin{tabular}{lp{4.5cm}p{10cm}}
 \toprule
 \textbf{Parameter} & \textbf{Description} & \textbf{Role in SAVER} \\ \midrule
 $\Delta\theta$ & Angular step size (degrees) & Defines the total number of discrete angles $K$. \\
 $\xi^2$ & Prior regularization parameter & Controls the variance of the Gaussian prior on the image $\mathbf{x}$. \\
 $T_{\text{initial}}$ & Initial temperature & Sets the initial level of randomness in the Softmax policy to promote global exploration. \\
$T_{\text{min}}$ & Minimum temperature & Defines the baseline temperature to maintain a degree of exploration even in the late stages. \\
$\eta$ & Annealing rate & Determines the speed of transition from broad exploration to focused exploitation of high-variance directions. \\
$N$ & Total acquisition rounds & Represents the total radiation dose budget (total number of X-ray projection rays). \\ \bottomrule
\end{tabular}
\end{table}



\begin{figure}[t]
\centering
\includegraphics[width=\linewidth]{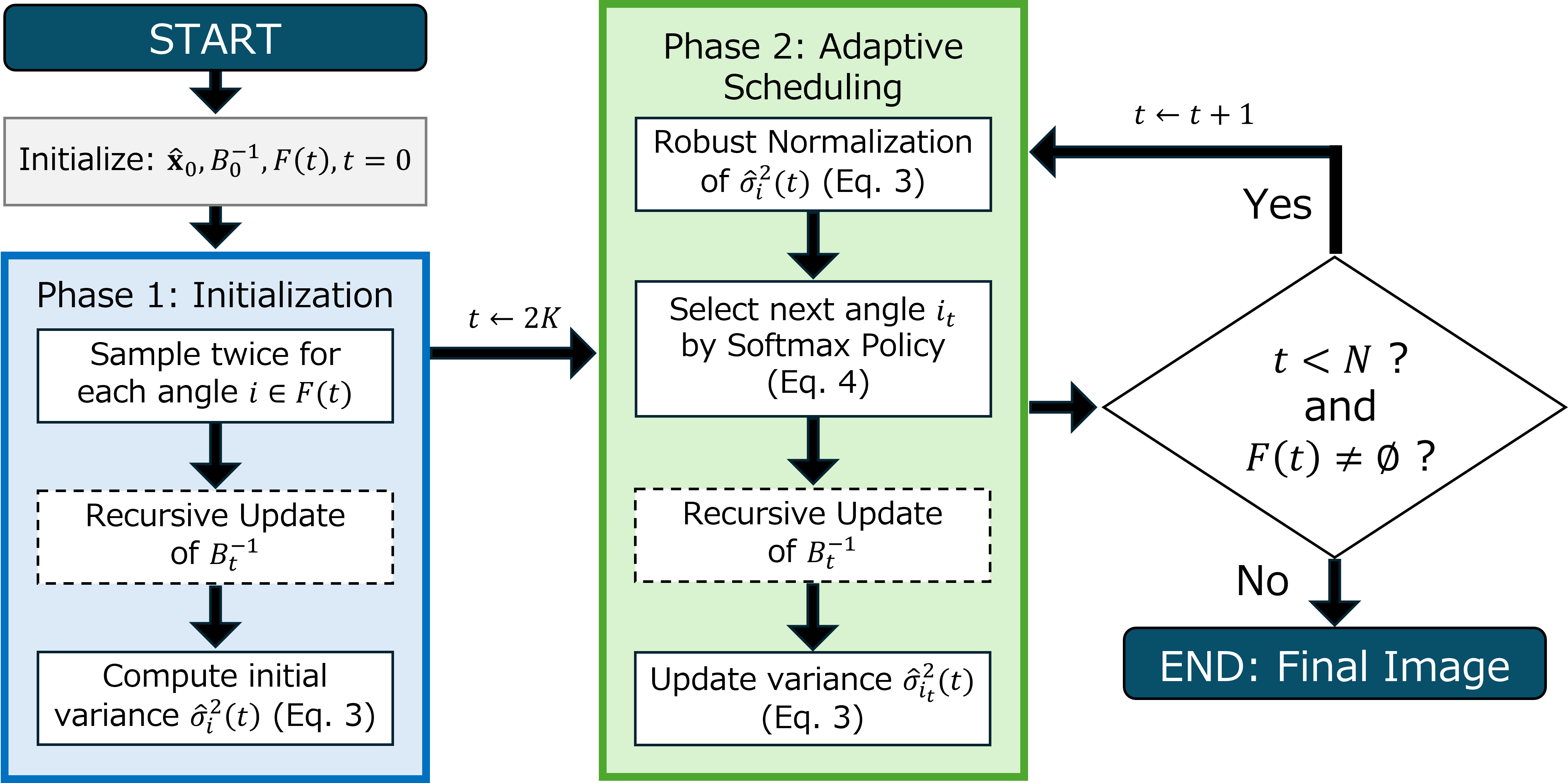}
\caption{\textbf{Operational workflow of the SAVER acquisition strategy}. \textcolor{black}{The framework consists of two main stages: (1) Phase 1 (Initialization), where two different \ns{projection data values} per angle are acquired to calculate initial sample variances for all potential directions; and (2) Phase 2 (Adaptive Scheduling), where the next projection angle is probabilistically selected via a Softmax-based policy with simulated annealing. 
In SAVER, it is not necessary to restore images in every round, so the process enclosed by the dashed line does not need to be performed. However, in the simulation experiment, in order to observe the progression of the evaluation of the restored images, the recursive update was conducted to restore the images in every round.
}}
\label{fig:proposed}
\end{figure}

\subsection*{Efficient Image Reconstruction via Woodbury Matrix Identity}
To achieve real-time adaptive acquisition, the image estimate $\hat{\mathbf{x}}_t$ {can} be updated at each round $t$ as {a new \ns{projection data value}} $y_t$ (Eq.~\ref{eq:projection}) is acquired. Solving the minimization problem in Eq.~\ref{eq:normal eq} directly requires inverting a $d \times d$ matrix \ns{at every round}, which entails a prohibitive computational cost of $O(d^3)$. In this study, we implement an efficient recursive update scheme based on the Woodbury matrix identity (also known as the matrix inversion lemma)~\cite{sherman1950adjustment} \tx{to estimate the reconstruction fidelity of SAVER and its variants along consecutive rounds.}
This approach allows to compute the updated inverse matrix $B_t^{-1} = (A_t^\top A_t + \frac{\sigma_B^2}{\xi^2}I)^{-1}$ using the result from the previous round $B_{t-1}^{-1}$, reducing the per-update complexity to $O(d^2)$. At each round $t$, given the selected index $i_t$ and irradiation position $r_t$, the estimate $\hat{\mathbf{x}}_t$ and the inverse covariance matrix $B_t^{-1}$ are updated as follows:

\begin{equation*}
\begin{cases}
D_t = 1 + \mathbf{a}_{i_t, r_t}^\top B_{t-1}^{-1} \mathbf{a}_{i_t, r_t} \\
\mathbf{k}_t = (B_{t-1}^{-1} \mathbf{a}_{i_t, r_t}) / D_t \\
\hat{\mathbf{x}}_t = \hat{\mathbf{x}}_{t-1} + \mathbf{k}_t (y_t - \mathbf{a}_{i_t, r_t}^\top \hat{\mathbf{x}}_{t-1}) \\
B_t^{-1} = B_{t-1}^{-1} - \mathbf{k}_t (\mathbf{a}_{i_t, r_t}^\top B_{t-1}^{-1})
\end{cases}
\end{equation*}

where $D_t$ is a scalar denominator and $\mathbf{k}_t \in \mathbb{R}^d$ is the gain vector. The recursion is initialized with $\hat{\mathbf{x}}_0 = \mathbf{0}$ and $B_0^{-1} = \frac{\xi^2}{\sigma_B^2} I$. This recursive formulation enables the SAVER framework to perform high-speed, sequential image updates, which is essential for real-time projection selection.
%
%

While the current recursive update using the Woodbury identity reduces the complexity to $O(d^2)$ per measurement, this remains computationally demanding for clinical resolutions (e.g., $512\times512$ pixels). In such high-dimensional spaces, the storage and update of the inverse covariance matrix $B_t^{-1}$ 
require significant memory and FLOPs. Future implementations \tk{may} leverage domain-decomposition methods, low-rank approximations of the covariance matrix, or fast iterative solvers to maintain the real-time adaptability of SAVER in large-scale settings. \tx{Nevertheless, as highlighted in the operational workflow in Figure~\ref{fig:proposed}, the full restoration of the image $\hat{\mathbf{x}}_t$ \tx{at every round is not mandatory}
for adaptive decision-making. Since the selection policy in SAVER is driven by the statistical variance of the observed \ns{projection data values}, the image reconstruction process can be deferred to the final stage or performed only when necessary, thereby significantly reducing the total computational overhead during the acquisition phase.}

\subsection*{Evaluation Metric}
\textcolor{black}{
To evaluate the reconstruction quality, we employ\tk{ed} the Structural Similarity Index Measure (SSIM)~\cite{wang2003multiscale}, which measures perceptual similarity by comparing luminance, contrast, and structural information between two images $\mathbf{x}$ and $\mathbf{y}$ \tk{$\in \mathbb{R}^{32} \times \mathbb{R}^{32}$}.
In practice, SSIM is computed as the mean SSIM (MSSIM) over local sliding windows. For each pixel location $(i,j)$, a local window $w_{i,j}$ of size $7 \times 7$ \tk{was} defined, and the local SSIM \tk{was} computed as $\mathrm{SSIM}_{i,j}(\mathbf{x},\mathbf{y})
=
\frac{(2\mu_{i,j}^\mathbf{x} \mu_{i,j}^\mathbf{y} + C_1)(2\sigma_{i,j}^\mathbf{xy} + C_2)}
{((\mu_{i,j}^\mathbf{x})^2 + (\mu_{i,j}^\mathbf{y})^2 + C_1)((\sigma_{i,j}^\mathbf{x})^2 + (\sigma_{i,j}^\mathbf{y})^2 + C_2)}$, where $\mu_{i,j}^\mathbf{x}$, $\mu_{i,j}^\mathbf{y}$ denote the local means, $\sigma_{i,j}^\mathbf{x}$, $\sigma_{i,j}^\mathbf{y}$ denote the local standard deviations, and $\sigma_{i,j}^\mathbf{xy}$ denotes the local covariance computed within the window $w_{i,j}$. Here, the constants $C_1$ and $C_2$ are defined as $C_1 = (K_1 L)^2, C_2 = (K_2 L)^2$, where $L$ denotes the dynamic range of the pixel intensities, and $K_1$ and $K_2$ are 
\tk{some} 
constants \tk{to avoid numerical instability that could arise for a very small divisor and we used} the standard values $K_1 = 0.01$ and $K_2 = 0.03.$~\cite{wang2003multiscale}}

\textcolor{black}{The overall similarity is obtained by averaging the local SSIM values over a specific set of pixel locations. In this study, the evaluation \tk{was} restricted to the inscribed circular region of the image domain. Let $\Omega_{\text{circle}}$ denote the set of pixel indices within this inscribed circle. The reconstruction quality at step $t$ is defined as the mean of the local SSIM indices within this region:
\begin{equation}\label{eq:ssim}
\mathrm{SSIM}(t)
=
\frac{1}{|\Omega_{\text{circle}}|}
\sum_{(i,j)\in\Omega_{\text{circle}}}
\mathrm{SSIM}_{i,j}(\hat{\mathbf{x}}_t,\mathbf{x}^*),
\end{equation} where $\hat{\mathbf{x}}_t$ denotes the reconstructed image at round $t$, $\mathbf{x}^*$ denotes the ground truth image.}

In implementation, we \tk{used} the \texttt{structural\_similarity} function from the \texttt{scikit-image.metrics} library {(version 0.25.2)} with its default settings~\cite{van2014scikit}, corresponding to a sliding window of size $7 \times 7$ and unit stride. In regions near the image boundaries where the local window extends beyond the image domain, reflect padding~\cite{liu2018partial} is employed, in which pixel values are symmetrically extended across the boundary. Compared to zero padding and other simple boundary treatments, this approach reduces distortion in local statistics (mean and variance), making it more suitable for stable SSIM evaluation.

\bibliography{sample}

\begin{thebibliography}{10}
\urlstyle{rm}
\expandafter\ifx\csname url\endcsname\relax
  \def\url#1{\texttt{#1}}\fi
\expandafter\ifx\csname urlprefix\endcsname\relax\def\urlprefix{URL }\fi
\expandafter\ifx\csname doiprefix\endcsname\relax\def\doiprefix{DOI: }\fi
\providecommand{\bibinfo}[2]{#2}
\providecommand{\eprint}[2][]{\url{#2}}

\bibitem{hounsfield1973computerized}
\bibinfo{author}{Hounsfield, G.~N.}
\newblock \bibinfo{journal}{\bibinfo{title}{Computerized transverse axial scanning (tomography): Part 1. description of system}}.
\newblock {\emph{\JournalTitle{The British journal of radiology}}} \textbf{\bibinfo{volume}{46}}, \bibinfo{pages}{1016--1022} (\bibinfo{year}{1973}).

\bibitem{brenner2007computed}
\bibinfo{author}{Brenner, D.~J.} \& \bibinfo{author}{Hall, E.~J.}
\newblock \bibinfo{journal}{\bibinfo{title}{Computed tomography—an increasing source of radiation exposure}}.
\newblock {\emph{\JournalTitle{New England journal of medicine}}} \textbf{\bibinfo{volume}{357}}, \bibinfo{pages}{2277--2284} (\bibinfo{year}{2007}).

\bibitem{valentin20082007}
\bibinfo{author}{Valentin, J.} \emph{et~al.}
\newblock \bibinfo{journal}{\bibinfo{title}{The 2007 recommendations of the international commission on radiological protection}}.
\newblock {\emph{\JournalTitle{ICRP publication}}} \textbf{\bibinfo{volume}{103}}, \bibinfo{pages}{2--4} (\bibinfo{year}{2008}).

\bibitem{beister2012iterative}
\bibinfo{author}{Beister, M.}, \bibinfo{author}{Kolditz, D.} \& \bibinfo{author}{Kalender, W.~A.}
\newblock \bibinfo{journal}{\bibinfo{title}{Iterative reconstruction methods in {X-ray CT}}}.
\newblock {\emph{\JournalTitle{Physica medica}}} \textbf{\bibinfo{volume}{28}}, \bibinfo{pages}{94--108} (\bibinfo{year}{2012}).

\bibitem{willemink2013iterative}
\bibinfo{author}{Willemink, M.~J.} \emph{et~al.}
\newblock \bibinfo{journal}{\bibinfo{title}{Iterative reconstruction techniques for computed tomography {Part} 1: technical principles}}.
\newblock {\emph{\JournalTitle{European radiology}}} \textbf{\bibinfo{volume}{23}}, \bibinfo{pages}{1623--1631} (\bibinfo{year}{2013}).

\bibitem{caiafa2013multidimensional}
\bibinfo{author}{Caiafa, C.~F.} \& \bibinfo{author}{Cichocki, A.}
\newblock \bibinfo{journal}{\bibinfo{title}{Multidimensional compressed sensing and their applications}}.
\newblock {\emph{\JournalTitle{Wiley Interdisciplinary Reviews: Data Mining and Knowledge Discovery}}} \textbf{\bibinfo{volume}{3}}, \bibinfo{pages}{355--380} (\bibinfo{year}{2013}).

\bibitem{kudo2013image}
\bibinfo{author}{Kudo, H.}, \bibinfo{author}{Suzuki, T.} \& \bibinfo{author}{Rashed, E.~A.}
\newblock \bibinfo{journal}{\bibinfo{title}{Image reconstruction for sparse-view {CT and interior CT—introduction to compressed sensing and differentiated backprojection}}}.
\newblock {\emph{\JournalTitle{Quantitative imaging in medicine and surgery}}} \textbf{\bibinfo{volume}{3}}, \bibinfo{pages}{147} (\bibinfo{year}{2013}).

\bibitem{chen2017low}
\bibinfo{author}{Chen, H.} \emph{et~al.}
\newblock \bibinfo{journal}{\bibinfo{title}{Low-dose {CT} via convolutional neural network}}.
\newblock {\emph{\JournalTitle{Biomedical optics express}}} \textbf{\bibinfo{volume}{8}}, \bibinfo{pages}{679--694} (\bibinfo{year}{2017}).

\bibitem{wolterink2017generative}
\bibinfo{author}{Wolterink, J.~M.}, \bibinfo{author}{Leiner, T.}, \bibinfo{author}{Viergever, M.~A.} \& \bibinfo{author}{I{\v{s}}gum, I.}
\newblock \bibinfo{journal}{\bibinfo{title}{Generative adversarial networks for noise reduction in low-dose {CT}}}.
\newblock {\emph{\JournalTitle{IEEE transactions on medical imaging}}} \textbf{\bibinfo{volume}{36}}, \bibinfo{pages}{2536--2545} (\bibinfo{year}{2017}).

\bibitem{dahmen2016feature}
\bibinfo{author}{Dahmen, T.} \emph{et~al.}
\newblock \bibinfo{journal}{\bibinfo{title}{Feature adaptive sampling for scanning electron microscopy}}.
\newblock {\emph{\JournalTitle{Scientific reports}}} \textbf{\bibinfo{volume}{6}}, \bibinfo{pages}{25350} (\bibinfo{year}{2016}).

\bibitem{andersen2012statistical}
\bibinfo{author}{Andersen, P.~K.}, \bibinfo{author}{Borgan, O.}, \bibinfo{author}{Gill, R.~D.} \& \bibinfo{author}{Keiding, N.}
\newblock \emph{\bibinfo{title}{Statistical models based on counting processes}} (\bibinfo{publisher}{Springer Science \& Business Media}, \bibinfo{year}{2012}).

\bibitem{kirkpatrick1983optimization}
\bibinfo{author}{Kirkpatrick, S.}, \bibinfo{author}{Gelatt~Jr, C.~D.} \& \bibinfo{author}{Vecchi, M.~P.}
\newblock \bibinfo{journal}{\bibinfo{title}{Optimization by simulated annealing}}.
\newblock {\emph{\JournalTitle{science}}} \textbf{\bibinfo{volume}{220}}, \bibinfo{pages}{671--680} (\bibinfo{year}{1983}).

\bibitem{sutton1998reinforcement}
\bibinfo{author}{Sutton, R.~S.}, \bibinfo{author}{Barto, A.~G.} \emph{et~al.}
\newblock \emph{\bibinfo{title}{Reinforcement learning: {An} introduction}}, vol.~\bibinfo{volume}{1} (\bibinfo{publisher}{MIT press Cambridge}, \bibinfo{year}{1998}).

\bibitem{jang2016categorical}
\bibinfo{author}{Jang, E.}, \bibinfo{author}{Gu, S.} \& \bibinfo{author}{Poole, B.}
\newblock \bibinfo{journal}{\bibinfo{title}{Categorical reparameterization with gumbel-softmax}}.
\newblock {\emph{\JournalTitle{arXiv preprint arXiv:1611.01144}}}  (\bibinfo{year}{2016}).

\bibitem{joseph2007improved}
\bibinfo{author}{Joseph, P.~M.}
\newblock \bibinfo{journal}{\bibinfo{title}{An improved algorithm for reprojecting rays through pixel images}}.
\newblock {\emph{\JournalTitle{IEEE transactions on medical imaging}}} \textbf{\bibinfo{volume}{1}}, \bibinfo{pages}{192--196} (\bibinfo{year}{2007}).

\bibitem{tikhonov1943stability}
\bibinfo{author}{Tikhonov, A.~N.} \emph{et~al.}
\newblock \bibinfo{title}{On the stability of inverse problems}.
\newblock In \emph{\bibinfo{booktitle}{Dokl. akad. nauk sssr}}, vol.~\bibinfo{volume}{39}, \bibinfo{pages}{195--198} (\bibinfo{year}{1943}).

\bibitem{shepp1974fourier}
\bibinfo{author}{Shepp, L.~A.} \& \bibinfo{author}{Logan, B.~F.}
\newblock \bibinfo{journal}{\bibinfo{title}{The {Fourier} reconstruction of a head section}}.
\newblock {\emph{\JournalTitle{IEEE Transactions on nuclear science}}} \textbf{\bibinfo{volume}{21}}, \bibinfo{pages}{21--43} (\bibinfo{year}{1974}).

\bibitem{joe2008constructing}
\bibinfo{author}{Joe, S.} \& \bibinfo{author}{Kuo, F.~Y.}
\newblock \bibinfo{journal}{\bibinfo{title}{Constructing {Sobol} sequences with better two-dimensional projections}}.
\newblock {\emph{\JournalTitle{SIAM Journal on Scientific Computing}}} \textbf{\bibinfo{volume}{30}}, \bibinfo{pages}{2635--2654} (\bibinfo{year}{2008}).

\bibitem{loh1996latin}
\bibinfo{author}{Loh, W.-L.}
\newblock \bibinfo{journal}{\bibinfo{title}{On {Latin} hypercube sampling}}.
\newblock {\emph{\JournalTitle{The annals of statistics}}} \textbf{\bibinfo{volume}{24}}, \bibinfo{pages}{2058--2080} (\bibinfo{year}{1996}).

\bibitem{parada2019blue}
\bibinfo{author}{Parada-Mayorga, A.}, \bibinfo{author}{Lau, D.~L.}, \bibinfo{author}{Giraldo, J.~H.} \& \bibinfo{author}{Arce, G.~R.}
\newblock \bibinfo{journal}{\bibinfo{title}{Blue-noise sampling on graphs}}.
\newblock {\emph{\JournalTitle{IEEE Transactions on Signal and Information Processing over Networks}}} \textbf{\bibinfo{volume}{5}}, \bibinfo{pages}{554--569} (\bibinfo{year}{2019}).

\bibitem{pukelsheim2006optimal}
\bibinfo{author}{Pukelsheim, F.}
\newblock \emph{\bibinfo{title}{Optimal design of experiments}} (\bibinfo{publisher}{SIAM}, \bibinfo{year}{2006}).

\bibitem{lattimore2020bandit}
\bibinfo{author}{Lattimore, T.} \& \bibinfo{author}{Szepesv{\'a}ri, C.}
\newblock \emph{\bibinfo{title}{Bandit algorithms}} (\bibinfo{publisher}{Cambridge University Press}, \bibinfo{year}{2020}).

\bibitem{auer2002finite}
\bibinfo{author}{Auer, P.}, \bibinfo{author}{Cesa-Bianchi, N.} \& \bibinfo{author}{Fischer, P.}
\newblock \bibinfo{journal}{\bibinfo{title}{Finite-time analysis of the multiarmed bandit problem}}.
\newblock {\emph{\JournalTitle{Machine learning}}} \textbf{\bibinfo{volume}{47}}, \bibinfo{pages}{235--256} (\bibinfo{year}{2002}).

\bibitem{hou2022almost}
\bibinfo{author}{Hou, Y.}, \bibinfo{author}{Tan, V.~Y.} \& \bibinfo{author}{Zhong, Z.}
\newblock \bibinfo{journal}{\bibinfo{title}{Almost optimal variance-constrained best arm identification}}.
\newblock {\emph{\JournalTitle{IEEE Transactions on Information Theory}}} \textbf{\bibinfo{volume}{69}}, \bibinfo{pages}{2603--2634} (\bibinfo{year}{2022}).

\bibitem{patel2015cone}
\bibinfo{author}{Patel, S.} \emph{et~al.}
\newblock \bibinfo{journal}{\bibinfo{title}{Cone beam computed tomography in e ndodontics--a review}}.
\newblock {\emph{\JournalTitle{International endodontic journal}}} \textbf{\bibinfo{volume}{48}}, \bibinfo{pages}{3--15} (\bibinfo{year}{2015}).

\bibitem{sherman1950adjustment}
\bibinfo{author}{Sherman, J.} \& \bibinfo{author}{Morrison, W.~J.}
\newblock \bibinfo{journal}{\bibinfo{title}{Adjustment of an inverse matrix corresponding to a change in one element of a given matrix}}.
\newblock {\emph{\JournalTitle{The Annals of Mathematical Statistics}}} \textbf{\bibinfo{volume}{21}}, \bibinfo{pages}{124--127} (\bibinfo{year}{1950}).

\bibitem{wang2003multiscale}
\bibinfo{author}{Wang, Z.}, \bibinfo{author}{Simoncelli, E.~P.} \& \bibinfo{author}{Bovik, A.~C.}
\newblock \bibinfo{title}{Multiscale structural similarity for image quality assessment}.
\newblock In \emph{\bibinfo{booktitle}{The thrity-seventh asilomar conference on signals, systems \& computers, 2003}}, vol.~\bibinfo{volume}{2}, \bibinfo{pages}{1398--1402} (\bibinfo{organization}{Ieee}, \bibinfo{year}{2003}).

\bibitem{van2014scikit}
\bibinfo{author}{Van~der Walt, S.} \emph{et~al.}
\newblock \bibinfo{journal}{\bibinfo{title}{scikit-image: image processing in {Python}}}.
\newblock {\emph{\JournalTitle{PeerJ}}} \textbf{\bibinfo{volume}{2}}, \bibinfo{pages}{e453} (\bibinfo{year}{2014}).

\bibitem{liu2018partial}
\bibinfo{author}{Liu, G.} \emph{et~al.}
\newblock \bibinfo{journal}{\bibinfo{title}{Partial convolution based padding}}.
\newblock {\emph{\JournalTitle{arXiv preprint arXiv:1811.11718}}}  (\bibinfo{year}{2018}).

\end{thebibliography}



\section*{Funding}
\tk{This work was partially supported by Technology Agency (JST) / Core Research for Evolutional Science and
Technology (CREST), Grant Number JPMJCR2333, Japan (to T.K. and J.H.), Grant Number JPMJCR2335 (to W.Y. and H.K.), and JST SPRING, Grant Number JPMJSP2119 (to S.N.). This work was also performed under the Cooperative Research Program of ``Network Joint Research Center for Materials and Devices (MEXT)''.  Institute for Chemical Reaction
Design and Discovery (ICReDD) was established by World Premier International Research
Initiative (WPI), MEXT, Japan.}

\section*{Author contributions statement}

\tk{
S.N., K.T., and T.K. conceived the experiment(s),  S.N. conducted the experiment(s) and analyzed the results under the supervision of K.T., J.H., H.K., W.Y., and T.K.  
 All authors reviewed the manuscript.}
\section*{Competing interests}

\tk{The authors declare no competing interests.}

\section*{Data availability} 

The datasets generated and/or analysed during the current study are available in the SAVER\_CT repository, \\https://github.com/nonagashunta/SAVER\_CT.

\end{document}


\raggedbottom
\maketitle
\thispagestyle{empty}
\vspace{-80pt} 

\section*{Supplementary Information Overview}

This Supplementary Information provides additional theoretical discussions and experimental results supporting the main conclusions of the present study.
First, we present a mathematical interpretation of the axis-prior initialization strategy employed in AIRS, SAVER-A, and SAVER-AO. In particular, we discuss how the initial acquisition of orthogonal projection directions contributes to stabilization of the posterior covariance matrix, improvement of the conditioning of the sequential inverse problem, and reduction of variance estimation instability during the earliest acquisition rounds.
We additionally discuss the numerical characteristics of the system matrix constructed using Joseph's \ns{reprojection} method~\cite{joseph2007improved}. Particular attention is paid to the relationship between projection orientation, \ns{reprojection}-induced smoothing effects, and the stability of early-stage sequential reconstruction.
The supplementary figures provide detailed convergence behaviors and statistical comparisons under a high-noise condition.
Specifically, we report:

\begin{itemize}
\item
SSIM convergence curves as a function of acquisition rounds {under a relatively higher noise condition ($\sigma_B=10^{-2}$)},
\item
distributions of normalized AUC values across independent trials.
\end{itemize}

These additional analyses further support the  {stability and} effectiveness  
of the proposed SAVER framework and its axis-prior extensions {within the investigated parameter ranges}.

\section{\tk{Why Axis-Prior Initialization is Effective in AIRS, SAVER-A, and SAVER-AO}}

In this supplementary section, we provide a mathematical interpretation of why the axis-prior initialization employed in AIRS, SAVER-A, and SAVER-AO contributes to efficient early-stage reconstruction in the proposed sequential CT acquisition framework.

\subsection*{Role of Early Measurements in Sequential Reconstruction}

Recall that the reconstruction problem is formulated as the following Tikhonov-regularized inverse problem~\cite{tikhonov1943stability}: $\hat{\mathbf{x}}_t=\left(A_t^\top A_t+\frac{\sigma_B^2}{\xi^2}I\right)^{-1}A_t^\top \mathbf{y}_t$, where $A_t \in \mathbb{R}^{t \times d}$ denotes the measurement matrix composed of sequentially acquired projection rays.
The corresponding posterior covariance matrix is $B^{-1}_t=\left(A_t^\top A_t+\frac{\sigma_B^2}{\xi^2}I\right)^{-1}$.
The quality of the reconstruction is therefore strongly governed by the conditioning and rank structure of $A_t^\top A_t$.
In the early acquisition stage ($t \ll d$), the matrix $A_t^\top A_t$ is highly rank-deficient, and thus the reconstruction quality depends critically on how efficiently the initial measurements span the image space.


In AIRS, SAVER-A, and SAVER-AO, all detector positions corresponding to \(0^\circ\) and \(90^\circ\) are first exhausted before entering the adaptive scheduling stage. Denote by $A_{\mathrm{axis}}=
\begin{pmatrix}
A_{0^\circ} \\
A_{90^\circ}
\end{pmatrix}$ the submatrix consisting only of projection rays from the horizontal and vertical directions.
For a Cartesian image grid, the rows of \(A_{0^\circ}\) approximately correspond to horizontal line sums, while the rows of \(A_{90^\circ}\) correspond to vertical line sums. Consequently, $A_{\mathrm{axis}}^\top A_{\mathrm{axis}}$ provides early constraints on both horizontal and vertical spatial frequencies of the image. This can be interpreted as constructing an initial low-frequency basis for the posterior estimate.
In contrast, purely random initialization may produce highly anisotropic sampling patterns \tk{especially} during early rounds, resulting in unstable posterior covariance structures.

\subsection*{Reduction of Posterior Uncertainty}

The recursive Woodbury update~\cite{sherman1950adjustment} employed in this study is 

\begin{equation*}
\begin{cases}
D_t = 1 + \mathbf{a}_{i_t, r_t}^\top B_{t-1}^{-1} \mathbf{a}_{i_t, r_t} \\
\mathbf{k}_t = (B_{t-1}^{-1} \mathbf{a}_{i_t, r_t}) / D_t \\
\hat{\mathbf{x}}_t = \hat{\mathbf{x}}_{t-1} + \mathbf{k}_t (y_t - \mathbf{a}_{i_t, r_t}^\top \hat{\mathbf{x}}_{t-1}) \\
B_t^{-1} = B_{t-1}^{-1} - \mathbf{k}_t (\mathbf{a}_{i_t, r_t}^\top B_{t-1}^{-1})
\end{cases}.
\end{equation*}
The reduction of posterior uncertainty caused by a newly-selected ray $\mathbf{a}_{i_t,r_t}$ is governed by $\mathbf{a}_{i_t,r_t}^\top
B^{-1}_{t-1}
\mathbf{a}_{i_t,r_t}$, which represents the uncertainty along the measurement direction.
When the initial rays are concentrated only on a limited subset of directions, the posterior covariance remains highly elongated in unobserved spatial directions. Axis-prior initialization mitigates this effect by rapidly reducing uncertainty along the two principal Cartesian directions.
As a result, the posterior covariance becomes more isotropic during early rounds, stabilizing subsequent adaptive scheduling.

\subsection*{Stabilization of Variance Estimation}

The proposed SAVER framework selects projection angles using the sample variance $\hat{\sigma}_i^2(t)=\frac{1}{n_i(t)}\sum_{y \in Y_i(t)}y^2-\left(\frac{1}{n_i(t)}\sum_{y \in Y_i(t)}y\right)^2.$ When \(n_i(t)\) is very small, the variance estimator becomes highly unstable: $\mathrm{Var}
\left[\hat{\sigma}_i^2(t)\right]=O\left(\frac{1}{n_i(t)}\right).$
Therefore, during the earliest acquisition rounds, inaccurate variance estimates may induce erroneous Softmax probabilities: $P_i(t)=\frac{\exp(\tilde{\sigma}_i(t)/T_t)}{\sum_j\exp(\tilde{\sigma}_j(t)/T_t)}.$ Axis-prior initialization functions as a deterministic warm-start procedure that stabilizes the initial posterior estimate before the adaptive variance-driven phase begins.
This \tk{can} 
reduce the probability that the scheduling policy becomes trapped in suboptimal projection directions caused by noisy early-stage variance estimates.



\subsection*{Relation Between AIRS, SAVER-A, and SAVER-AO}

\tk{
The three methods differ from each other on how the method selects the irradiation angles after the axis-prior initialization. 
After the initialization, AIRS performs uniform random sampling while
SAVER-A utilizes adaptive variance-driven scheduling that prioritizes exploration at earlier rounds by estimating underlying variance of \ns{projection data values} along each angle by the sample variances
and gradually shifts exploitation as the number of rounds passes. SAVER-AO uses axis-prior initialization followed by oracle-guided Softmax scheduling using the true variances.
Thus, AIRS isolates the contribution of the deterministic initialization itself, while SAVER-A and SAVER-AO additionally exploit adaptive concentration toward informative projection angles. 
The empirical gap between AIRS and SAVER-A therefore directly quantifies the utility of real-time variance estimation beyond simple axis-prior stabilization, and also a higher performance of SAVER-AO than SAVER-A observed for phantoms especially 
\tm{ {\it 3. Triangle} and {\it 7. Gradient}}  in Fig.~5 indicates that tighter estimation of the \ns{projection data values}' variance could result in better reconstruction performance.  
}

\section{Reconstruction Performance {under Relatively Higher Noise}}

{This section evaluates the behavior
\tk{of the proposed SAVER framework on a higher noise level $\sigma_B$ ($10^4$ times larger) than that presented in the main text.}
\tk{Here,}
we present \tk{corresponding} results for the representative setting ($\sigma_B = 10^{-2}, \eta = 0.01$) in 
Figures~\labelcref{fig:S1,fig:S2,fig:S3}.}


As shown in 
Figures~\ref{fig:S1} and~\ref{fig:S2}, SAVER-based methods outperform its non-adaptive baseline (c.f., SAVER and SAVER-O against Random, and SAVER-A and SAVER-AO against AIRS) for phantoms with high structural anisotropy (e.g., {\it 1. Rectangle}, 
{\it 4. Cross}, and {\it 6. Stripes})
or show comparable performance for most of the other phantoms with more homogeneous information distributions 
as seen in the lower noise case. For latter phantoms, SAVER-A exhibits performance comparable to AIRS, indicating that the deterministic axis-prior provides the primary contribution to reconstruction stability in such cases. \tm{A higher performance of SAVER-AO than SAVER-A was also observed for \tm{ {\it 3. Triangle} and {\it 7. Gradient}}, which suggests 
some improvement if a tighter estimation of the \ns{projection data values}' variance is possible as seen in a lower noise level.}

Figure~\ref{fig:S3} highlights the specific performance gap between random sampling, deterministic axis-prior initialization, and our adaptive scheduling. These results confirm that the performance trends observed in the low-noise environment (main text) are 
preserved even when the measurement noise is increased by four orders of magnitude ($\sigma_B=10^{-2}$). The synergy between orthogonal initial measurements and variance-driven exploration continues to yield efficient reconstruction, particularly for structured objects, demonstrating the stability of the SAVER-A framework against stochastic perturbations.

\subsection*{\tm{Sensitivity Analysis on Annealing Hyperparameter}}

We also conducted extensive evaluations by varying the annealing parameter $\eta = 0.1, 1.0,$ and $10$. In these settings, SAVER-A maintained its superiority over the other methods that do not use the information of true variance. Axis-prior initialization during the earliest acquisition rounds ($t < 100$) was seen against non-adaptive baseline Random. This suggests that the axis-prior serves as a stability warm-start mechanism across the tested signal-to-noise ratios and hyperparameter settings. 
Figures \ref{fig:eta01}, \ref{fig:eta1}, and \ref{fig:eta10} exemplified the SSIM convergence with the noise level 
$\sigma_B = 
10^{-6}$ for four methods, Random, AIRS, SAVER-A, and SAVER-AO.  
    The performance gap among SAVER-A, SAVER-AO, and AIRS across different $\eta$ values indicates that variance-driven exploration provides a meaningful advantage for anisotropic samples, although this gain is naturally moderated in isotropic cases: e.g., compare the isotropic phantoms such as {\it 5. Checkerboard} and {\it Shepp-Logan} and the other aisotropic phantoms over $\eta$s. For dependency on annealing parameter $\eta$ of SAVER-A performance, it should be noted that SAVER-A gets closer to the oracle SAVER-AO for some of the phantoms such as {\it 3. Triangle} and {\it 7. Gradient} as $\eta$ increases from $\eta=0.01$ to $\eta=0.1$ and $1.0$ in latter rounds $> 300$ although extremely high annealing rate $\eta=10$ suppresses this improvement. This may indicate that while most phantoms in this paper have less dependency on the annealing rates, there exists room to optimize the annealing schedule in practical applications. This is one of the forthcoming subjects to be unresolved. 
\begin{figure}[t]
\centering
\includegraphics[width=\linewidth]{figures/experiment_results/-2/0.01/ssim_curves_0.01_0.01_10.png}
\caption{\textbf{SSIM convergence for all methods ($\sigma_B = 10^{-2}, \eta = 0.01$).} Shaded regions indicate standard deviation.}
\label{fig:S1}
\end{figure}

\begin{figure}[t]
\centering
\includegraphics[width=\linewidth]{figures/experiment_results/-2/0.01/auc_boxplots_0.01_0.01_10.png}
\caption{\textbf{Distribution of AUC/500 values ($\sigma_B = 10^{-2}, \eta = 0.01$).} Boxplots represent 10 independent trials for each method.}
\label{fig:S2}
\end{figure}

\begin{figure}[t]
\centering
\includegraphics[width=\linewidth]{figures/experiment_results/-2/0.01/ssim_curves_axis_0.01_0.01_10.png}
\caption{{\textbf{Detailed SSIM convergence comparison of representative methods} ($\sigma_B = 10^{-2}, \eta = 0.01$). This figure highlights the performance gap between purely random sampling (Random), axis-prior initialization (AIRS), and the proposed adaptive variance-driven scheduling (SAVER-A/AO). Shaded regions indicate the standard deviation across 10 independent trials.}}
\label{fig:S3}
\end{figure}

\begin{figure}[t]
\centering
\includegraphics[width=\linewidth]{figures/experiment_results/-6/0.1/ssim_curves_axis_1e-06_0.1_10.png}
\caption{{\textbf{Detailed SSIM convergence comparison of representative methods} ($\sigma_B = 10^{-6}, \eta = 0.1$).}}
\label{fig:eta01}
\end{figure}

\begin{figure}[t]
\centering
\includegraphics[width=\linewidth]{figures/experiment_results/-6/1/ssim_curves_axis_1e-06_1_10.png}
\caption{{\textbf{Detailed SSIM convergence comparison of representative methods} ($\sigma_B = 10^{-6}, \eta = 1$).}}
\label{fig:eta1}
\end{figure}

\begin{figure}[t]
\centering
\includegraphics[width=\linewidth]{figures/experiment_results/-6/10/ssim_curves_axis_1e-06_10_10.png}
\caption{{\textbf{Detailed SSIM convergence comparison of representative methods} ($\sigma_B = 10^{-6}, \eta = 10$).}}
\label{fig:eta10}
\end{figure}

\bibliography{sample_si}